%% file: main.tex
\documentclass[conference]{IEEEtran}
\IEEEoverridecommandlockouts
\usepackage{cite}
\usepackage{amsmath,amssymb,amsfonts}
\usepackage{algorithmic}
\usepackage{graphicx}
\usepackage{textcomp}
\usepackage{multicol, multirow}
\usepackage{xcolor}
\usepackage{hyperref}
\usepackage{subcaption}
\usepackage{graphicx} % for \includegraphics
\usepackage{booktabs} % for better looking tables
\usepackage{array}
\usepackage{geometry}
\def\BibTeX{{\rm B\kern-.05em{\sc i\kern-.025em b}\kern-.08em
    T\kern-.1667em\lower.7ex\hbox{E}\kern-.125emX}}
\begin{document}

\title{Quizzard@INOVA Challenge 2025 - Track A: Plug-and-Play Technique in \\ Interleaved Multi-Image Model}

\author{\IEEEauthorblockN{Dinh Viet Cuong$^{1, \dagger}$, Hoang-Bao Le$^{1,2, \dagger}$ \thanks{$\dagger$ These authors have equal contribution.}, An Pham Ngoc Nguyen$^{1,2}$, Liting Zhou$^{1,2}$, Cathal Gurrin$^{1,2}$}
\IEEEauthorblockA{$^1$\textit{School of Computing, Dublin City University, Ireland}\\
\textit{$^{2}$ADAPT Centre, Ireland}\\
bao.le2@mail.dcu.ie, an.nguyen@adaptcentre.ie \{dinhviet.cuong, liting.zhou, cathal.gurrin\}@dcu.ie}}

\maketitle

\input{section/0-abstract}


% \footnote{(*)}

\input{section/1-introduction}

\input{section/2-approach}

% \section{Tasks and Datasets} \label{sec:2}

% While MMCoQA contains 2,382 samples, the other datasets all have 500 samples in the test set. 

% Lastly, the task Multi-Image Discrimination uses datasets like LFW for facial comparison and Totally-Looks-Like for evaluating visual similarity selection across multiple dimensions. 

\input{section/3-experiments}

\input{section/4-conclusion}

\section*{Acknowledgment}

    This publication has emanated from research supported in part by research grants from Science Foundation Ireland (SFI) under grant numbers  SFI/13/RC/2106\_P2 and 18/CRT/6223, and co-funded by the European Regional Development Fund.

\bibliographystyle{IEEEbib}
\bibliography{myref}

\end{document}

%% file: section/0-abstract.tex
\begin{abstract}
    This paper addresses two main objectives. Firstly, we demonstrate the impressive performance of the LLaVA-NeXT-interleave on 22 datasets across three different tasks: Multi-Image Reasoning, Documents and Knowledge-Based Understanding and Interactive Multi-Modal Communication. Secondly, we add the Dense Channel Integration (DCI) connector to the LLaVA-NeXT-Interleave and compare its performance against the standard model. We find that the standard model achieves the highest overall accuracy, excelling in vision-heavy tasks like VISION, NLVR2, and Fashion200K. Meanwhile, the DCI-enhanced version shows particular strength on datasets requiring deeper semantic coherence or structured change understanding such as MIT-States\_PropertyCoherence and SlideVQA. Our results highlight the potential of combining powerful foundation models with plug-and-play techniques for Interleave tasks. The code is available at \href{https://github.com/dinhvietcuong1996/icme25-inova}{https://github.com/dinhvietcuong1996/icme25-inova}. 
\end{abstract}

%% file: section/1-introduction.tex
\section{Introduction}

Transformer architectures have rapidly evolved from their initial success in machine translation~\cite{vaswani2017attention} to becoming the foundation of modern models like BERT~\cite{devlin2019bert}, GPT~\cite{brown2020gpt}, and T5~\cite{radford2021t5}. Beyond text, Transformers have been effectively applied to vision through Vision Transformers (ViTs)~\cite{dosovitskiy2020image}, which treat images as sequences of patches and outperform CNNs when scaled. As models matured, they enabled new tasks previously constrained by memory, including long-document classification~\cite{park2022efficient}, video understanding~\cite{bertasius2021space}, and large-scale vision-language learning. Notably, CLIP~\cite{radford2021clip} learns from 400M image-text pairs via contrastive learning, achieving strong zero-shot performance and cross-domain generalization.

Recent advancements in multimodal large language models (MLLMs) have seen the LLaVA \cite{liu2023improvedllava} framework evolve into specialized and optimized variants tailored for diverse applications. Core innovations include LLaVA-NeXT-Interleave \cite{liu2024llavanext}, which extends multimodal reasoning to interleaved image-video-3D contexts via dynamic resolution encoding, and LLaVA-Med \cite{li2023llavamed}, a biomedical VQA model trained with curriculum learning for robust pathology analysis. Efficiency-driven adaptations like LLaVA-UHD \cite{guo2024llava-uhd} address high-resolution image processing through modular slicing, while LLaVA-Mini \cite{llavamini} reduces computational overhead drastically by encoding images into single tokens. Parallel enhancements further optimize performance: LoGra-Med \cite{nguyen2024logramedlongcontextmultigraph} aligns medical modalities via graph-based triplet learning, KG-LLaVA \cite{hamza2025kgllava} integrates knowledge graphs for interpretable thoracic diagnostics, and plug-and-play techniques like Dense Connector \cite{yao2024dense} and LLaVA-PruMerge \cite{shang2024llavaprumerge} refine token efficiency for high-resolution or video inputs. Together, these innovations expand LLaVA’s applicability across precision-critical domains while balancing computational scalability and task-specific accuracy.

In this paper, we compare LLaVA-NeXT-Interleave finetuned with and without a dense connector and discuss the potential of the Plug-and-Play technique on three tasks of ICME25 Grand Challenge Inova's Track A: Multi-Image Reasoning, Document and Knowledge-Based Understanding, and Interactive Multi-Modal Communication.

%% file: section/2-approach.tex
\section{Approach} 

\begin{figure}[!h]
    \centering
    \includegraphics[width=\linewidth]{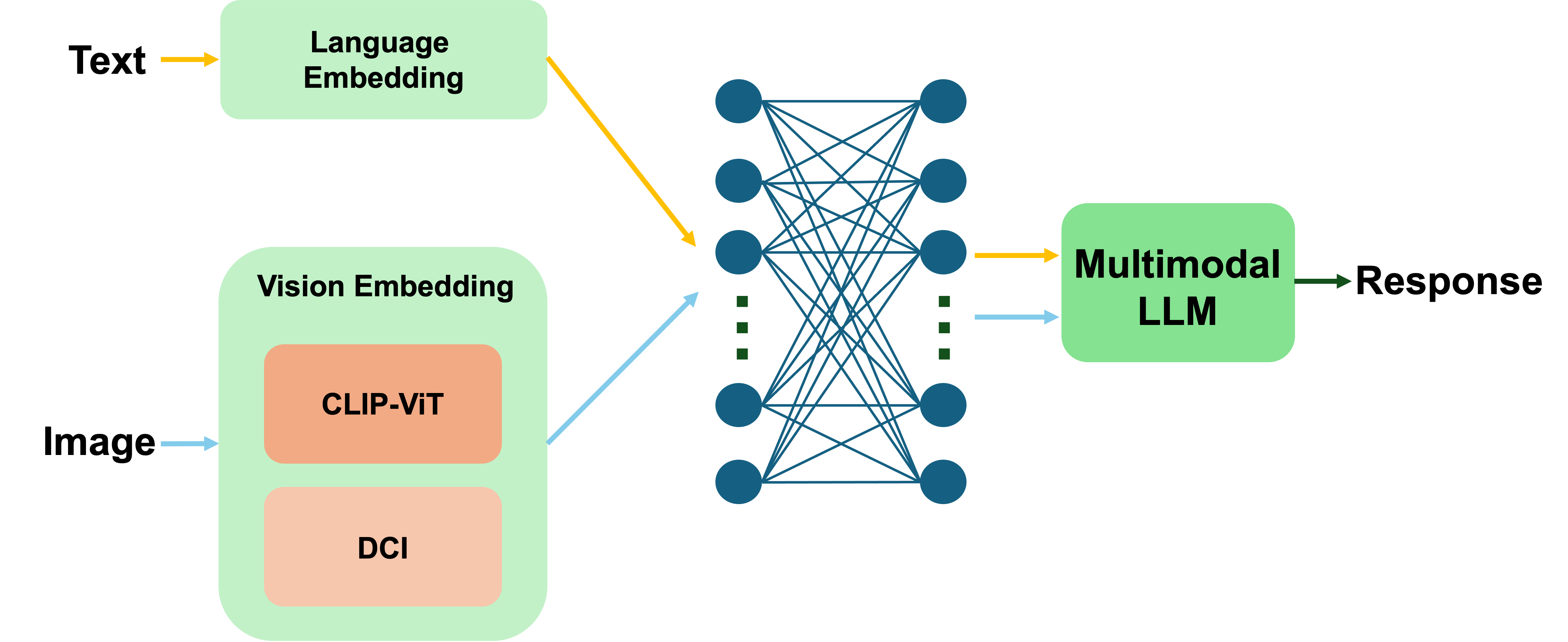}
    \caption{LLaVa-NeXT-Interleave with DCI Connector}
    \label{fig:full_model}
\end{figure}

\begin{figure}[!h]
    \centering
    \includegraphics[width=\linewidth]{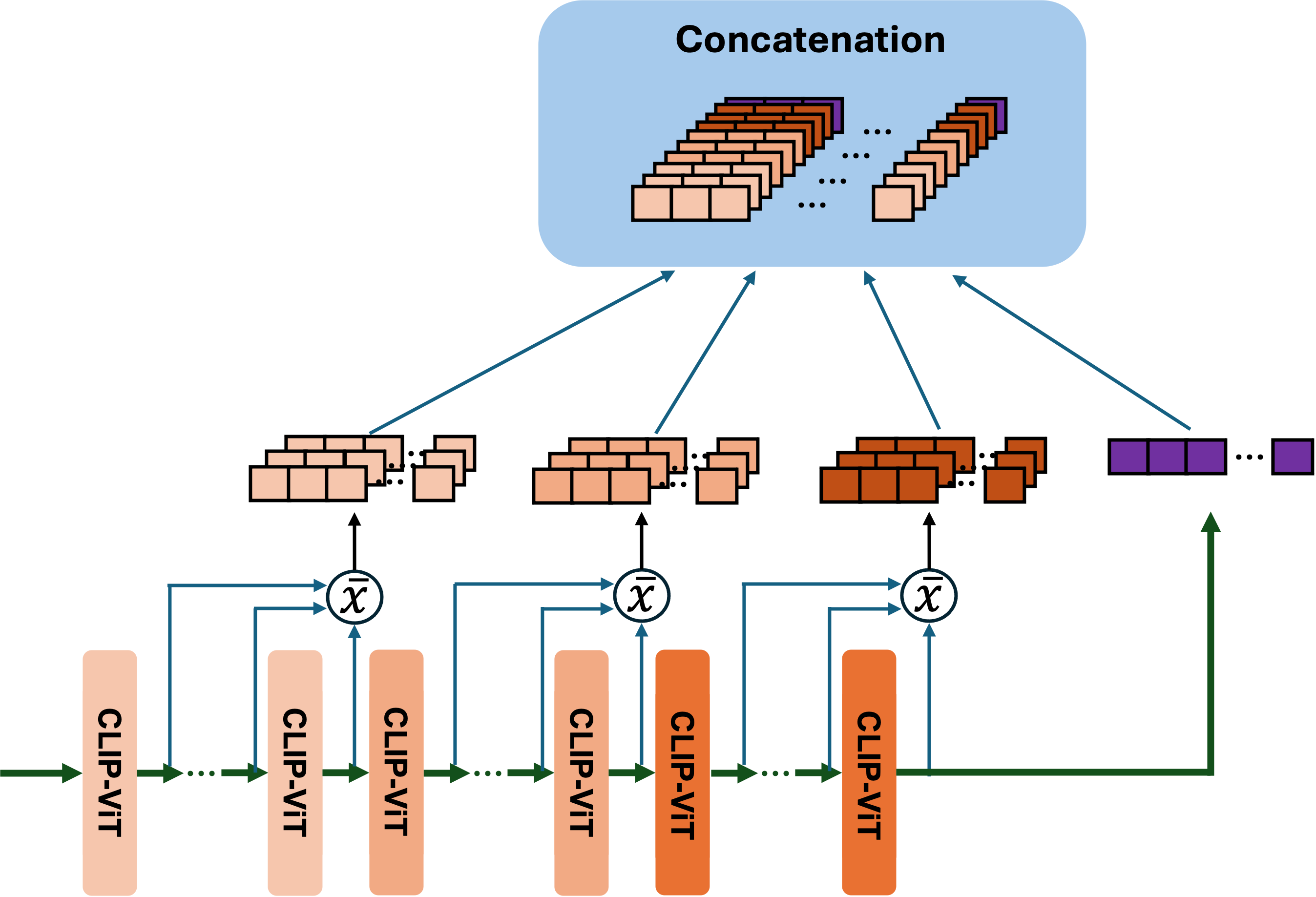}
    \caption{Vision Embedding block: CLIP-ViT-based Vision Encoder with Dense Channel Integration (DCI) Connector}
    \label{fig:dci}
\end{figure}

Given the outstanding performance of the LLaVA-NeXT-Interleave architecture\cite{liu2024llavanext} on interleaved images-texts tasks in the existing literature, we adopt and modify this model for the three aforementioned challenges. Specifically, we enhance the performance of the original LLaVa-Next-Interleave model to adapt with these specific tasks by adding a Dense Channel Integration (DCI) connector~\cite{yao2024dense} to the model. The description of the original LLaVa-NeXT-Interleave model is shown in Subsection \ref{sec:3.1}, the description of the DCI connector and how to integrate the dense connector into the LLaVa-NeXT-Interleave model are shown in Subsection \ref{sec:3.2}. To this end, the full modified architecture is shown in Figures \ref{fig:full_model}-\ref{fig:dci}.

\subsection{LLaVA-NeXT-Interleave Architecture} \label{sec:3.1}
The LLaVA-NeXT-Interleave model is built upon LLaVA-1.5 \cite{liu2023improvedllava}, which is an enhanced multimodal model that simplifies the architecture of its predecessor LLaVA \cite{liu2023llava} while achieving state-of-the-art performance across a range of vision-language tasks. Therefore, it can be said that LLaVA-NeXT-Interleave is the third generation of LLaVa. 

LLaVA was originally proposed as a vision-language model that connects a powerful large language model (LLM), such as Vicuna, with a vision encoder like CLIP-ViT, enabling it to process and understand multimodal inputs. The core design includes a visual encoder to extract image features, a projection layer to align these with the LLM’s embedding space, and the LLM itself for reasoning and generation. Building upon this foundation, LLaVA-1.5 presents an improved baseline by simplifying and optimizing the architecture. Notably, it replaces the original linear projection with a two-layer MLP, offering more expressive visual-language alignment. It retains the same high-resolution CLIP ViT-L/336px encoder and Vicuna-13B language model, but achieves superior performance using only 1.2 million publicly available image-text pairs. LLaVA-1.5 demonstrates that with effective instruction tuning and architectural refinement, high-quality multimodal understanding can be achieved with relatively low data and compute costs, surpassing prior baselines across a range of visual QA benchmarks.

LLaVA-NeXT-Interleave is a state-of-the-art open-source large multimodal model with improvements in reasoning, OCR capabilities, and world knowledge understanding compared to previous versions. While it retains the core architecture of combining a vision encoder, a projection module, and a large language model, LLaVA-NeXT-Interleave introduces several important improvements. In particular, by increasing the input image resolution to $4\times$ more pixels  with  three aspect ratios supported, including to $672\times 672$, $336\times 1344$ and $1344\times 336$, it captures much finer visual details across diverse image formats. Apart from that, it also introduces a more diverse and improved visual instruction tuning dataset, which boosts its performance in visual reasoning and OCR tasks. Furthermore, with enhanced instruction tuning and pretraining strategies, the model demonstrates stronger visual conversation abilities, adapting to a wider range of real-world scenarios with better world knowledge and logical reasoning. Lastly, efficient deployment and fast inference are enabled through SGLang, making LLaVA-NeXT-Interleave both powerful and practical for real-world applications.

\subsection{Dense Channel Integration Connector (DCI Connector)}\label{sec:3.2}
The Dense Channel Integration connector enhances visual perception in multimodal large language models (MLLMs) by processing and synthesizing visual features of all layers in the vision encoder instead of only outputting the final layer of the vision encoder like traditional methods. In this study, we add this dense connector to the original vision encoder of the LLaVA-NeXT-Interleave model. Therefore, the vision embedding output incorporates visual features from all layers. The resulting embedding then follows the same process as in the original LLaVA-NeXT-Interleave model, passing through an MLP layer. This technique addresses challenges of feature redundancy and higher dimensionality by ensuring dense connectivity across a wider range of visual layers without causing excessively high dimensions. Furthermore, it mitigates the issues that arise from simply concatenating all visual feature layers, which can be problematic during the training process.

Specifically, DCI partitions the features from $L$ layers of the vision encoder $V_i, 1 \leq i \leq L$ ($L$ CLIP-ViT layers in our case) into $G$ groups, with each group containing $M$ adjacent visual features where $M = \frac{L}{G}$. The features within group $i$ are averaged to yield $G$ fused visual representations, denoted as $GL_{i}$. The formula for calculating a fused visual representation $GL_{i}$ is as follows:
\begin{equation}
    GL_i = \frac{1}{M} \sum_{k = (i-1)M+1}^{iM} V_i, \hspace{1cm} 1 \leq i \leq G
\end{equation}
Finally, the vision embedding $VE$ is obtained by concatenating these fused visual representations and the final layer's visual feature along the channel (layer) dimension.
\begin{equation}
    \operatorname{EV} =\operatorname{Concatenate}([GL_1, GL_2, \cdots, GL_G, V_L])
\end{equation}

This dense connector is compatible with a variety of existing vision encoders (such as CLIP \cite{radford2021clip} and SigLIP \cite{zhai2023siglip}) and language models (including LLaMA \cite{touvron2023llama} and Yi \cite{young2024yi}), and introduces less than 2\% additional trainable parameters compared to baseline models. Furthermore, it streamlines the training process by requiring only single-stage instruction tuning, eliminating the need for multi-phase pretraining. Experimental results demonstrate that this approach yields significant performance gains-such as a 9.7\% improvement on visual question answering benchmarks-while preserving computational efficiency. The dense connectivity and hierarchical fusion enable the model to better capture spatial relationships and object details, addressing key limitations of previous sparse connector designs.

%% file: section/3-experiments.tex
\section{Tasks, Data and Evaluation Metrics}

\subsection{Tasks}
In this study, we focus on understanding visual content, aligning it with textual cues and performing logical inference to answer open-ended or multiple-choice questions that are relevant to multiple images, including:

\begin{itemize}
    \item \textbf{Multi-Image Reasoning}: this involves comparing and analyzing multiple images and their related texts to draw meaningful conclusions or highlight key differences.\\

    \item \textbf{Document and Knowledge-Based Understanding}: this involves understanding and extracting information from structured contents such as documents, charts, and tables.\\

    \item \textbf{Interactive Multi-Modal Communication}: this involves engaging in a conversation by understanding and responding to both images and text in a dynamic, interactive setting.\\

    % \item \textbf{Multi-Image Discrimination}: this evaluating two images side-by-side to identify both similarities and differences across various aspects like objects, actions, and environments.
\end{itemize}

\subsection{Datasets}

To train and evaluate our proposed model on the aforementioned tasks, we utilize 22 data sources. For Multi-Image Reasoning, datasets like Spot-the-Diff and CLEVR-Change are used for detecting visual differences in surveillance and synthetic environments, while IEdit and Birds-to-Words emphasize expressing nuanced visual relationships and subtle distinctions. Domains like driving, fashion and industrial scenarios are covered via nuScenes, Fashion200K and VISION, respectively. The MIT-States-PropertyCoherence dataset plays a critical role in assessing how well models can understand and evaluate the logical consistency of object properties across different states. For the second task, Document and Knowledge-Based Understanding, datasets include SlideVQA, OCR-VQA, WebQA, TQA, MMQA and DocVQA are used to train models' comprehension of slides, book covers, documents and webpages. Interactive Multi-Modal Communication task uses conversational datasets like ALFRED and MMCoQA, which challenge models to engage in coherent visual-language dialogue.

\subsection{Evaluation Metrics}
To evaluate our selected models in this study, we adopt two metrics as follows:
\begin{itemize}
    \item \textbf{ROUGE-L}: Used for evaluating the performance of the models on 6 datasets, namely Spot-the-Diff, CLEVR-Change, IEdit, Birds-to-Words, ALFRED and MMCoQA.
    \item \textbf{Accuracy}: Used for evaluating the performance of the models on the remaining datasets.
\end{itemize}
Additionally, we compute the average score within each metric group to better observe the differences in performance across model variants.

\section{Experiments and Results}
\subsection{Implementation Details}

For training and inference, we split the provided dataset into training and validation sets using a 9:1 ratio.

We follow the training and evaluation setup of \textbf{LLaVA-NeXT-Interleave}~\cite{liu2024llavanext}, using \textbf{SigLIP}~\cite{zhai2023siglip} as the vision backbone and the Qwen-based LLaVA-NeXT-Interleave model\footnote{\url{https://huggingface.co/lmms-lab/llava-next-interleave-qwen-7b}} with 7B parameters. The provided training set is split into training and validation subsets with a 9:1 ratio. Training is performed for 1 epoch using the Adam optimizer with a learning rate of $2 \times 10^{-5}$. For the DCI-enhanced version, we follow the implementation from~\cite{yao2024dense}. All experiments are run on one A100-80GB GPU.

\subsection{Results}

We report the results of three settings of LLaVA-NeXT-Interleave evaluated on the validation set. 

\paragraph{Performance Comparison}

\begin{table*}[!h]
\centering
\caption{Results on Track A's Tasks. We report the ROUGE-L score for the Interactive Multi-Modal Communication task and the Accuracy for the other two tasks.}
\begin{tabular}{|c|l|ccc|}
\hline
\multirow{2}{*}{\textbf{TASK}}  & \multicolumn{1}{c|}{\multirow{2}{*}{\textbf{DATASET}}} &  \multicolumn{3}{c|}{\textbf{MODEL}}    \\ \cline{3-5} 
& \multicolumn{1}{c|}{}      & FT w/ DCI   & FT  & Original \\ \hline\hline
\multirow{15}{*}{\textbf{Multi-Image Reasoning}}  & Spot-the-Diff     & 33.71&  34.16&  34.57    \\ % \cline{2-5}
& CLEVR-Change      & 59.26&  58.96&  50.54    \\ % \cline{2-5}
& IEdit      & 31.33&  34.35&  35.51    \\ % \cline{2-5} 
& Birds-to-Words    & 35.78 & 35.98&  36.12    \\ % \cline{2-5} 
& \textbf{AVERAGE (ROGUE-L)} & 40.02 & \textbf{40.86} & 39.19 \\ \cline{2-5}
& nuscenes   & 74.22&  78.05&  78.91    \\ % \cline{2-5}
& VISION     & 88.90&  88.55&  86.40    \\ % \cline{2-5} 
& Fashion200K& 76.55&  81.63&  41.15    \\ % \cline{2-5} 
& MIT-States\_PropertyCoherence & 91.50&  94.75&  93.50   \\ % \cline{2-5} 
& MIT-States\_StateCoherence & 98.75&  96.25&  84.00    \\ % \cline{2-5} 
& RecipeQA\_ImageCoherence   & 95.06&  98.24&  88.18    \\ % \cline{2-5} 
& NLVR2      & 81.84&  86.93&  83.64    \\ % \cline{2-5} 
& VizWiz     & 64.80&  64.10&  65.60     \\ % \cline{2-5} 
& \textbf{AVERAGE (Accuracy)} & 83.95 & \textbf{86.06} & 77.71  \\ 

\hline\hline
\multirow{7}{*}{\textbf{Document and Knowledge-Based Understanding}} &  SlideVQA   & 51.50&  65.25&  63.50    \\ % \cline{2-5} 
& OCR-VQA    & 77.50&  90.25&  90.50    \\ % \cline{2-5}
& WebQA      & 21.29&  22.83&  26.38    \\ % \cline{2-5} 
&  TQA        & 65.68&  75.54&  78.50    \\ % \cline{2-5}
& MultiModalQA     & 35.37&  36.46&  17.90    \\ % \cline{2-5} 
& ManyModalQA   & 43.70&  46.26&  11.52    \\ % \cline{2-5} 

& \textbf{AVERAGE (Accuracy)} & 49.17 & \textbf{56.10}  & 35.30 \\ 
\hline\hline

\multirow{2}{*}{\textbf{Interactive Multi-Modal  Communication}} & MMCoQA     & 60.35 & 64.93&  29.81    \\ % \cline{2-5} 
& ALFRED     & 67.89&  68.05&  66.91    \\ % \cline{2-5} 
 & \textbf{AVERAGE (ROGUE-L)} & 64.12 & \textbf{66.49} & 48.36 \\ 

\hline
 
\end{tabular}
\label{tab:results}
\end{table*}

The performance of three models used in this study is displayed in Table \ref{tab:results}. Specifically, fine-tuning the \textbf{LLaVA-NeXT-Interleave} model results in substantial performance improvements across most tasks. The fine-tuned version (FT) consistently outperforms the original, especially in \textbf{accuracy-driven benchmarks} such as Multi-Image Reasoning and Document Understanding, demonstrating the model’s adaptability when exposed to task-specific data.

Comparing the two fine-tuning strategies—\textbf{standard fine-tuning} and \textbf{DCI-enhanced fine-tuning (FT w/ DCI)}—reveals a trade-off. The standard FT model achieves the \textbf{highest overall accuracy}, excelling in vision-heavy tasks like \textit{VISION}, \textit{NLVR2}, and \textit{Fashion200K}. Meanwhile, the DCI-based version shows particular strength on datasets requiring deeper semantic coherence or structured change understanding, such as \textit{MIT-States\_PropertyCoherence} and \textit{SlideVQA}.

In the \textbf{Interactive Multi-Modal Communication} tasks, both fine-tuned variants significantly outperform the original model. However, the standard FT model slightly surpasses the DCI version, suggesting that while DCI improves grounding and coherence, it may not consistently benefit instruction-following or dialogue-heavy benchmarks.

In summary, DCI contributes valuable improvements in specific tasks involving \textbf{semantic alignment and coherence}, but the standard fine-tuning strategy remains more balanced and effective overall. These findings suggest a promising future direction in combining both strategies for \textbf{hybrid fine-tuning approaches}.

\begin{figure}[!h]
    \centering
    \includegraphics[width=\linewidth]{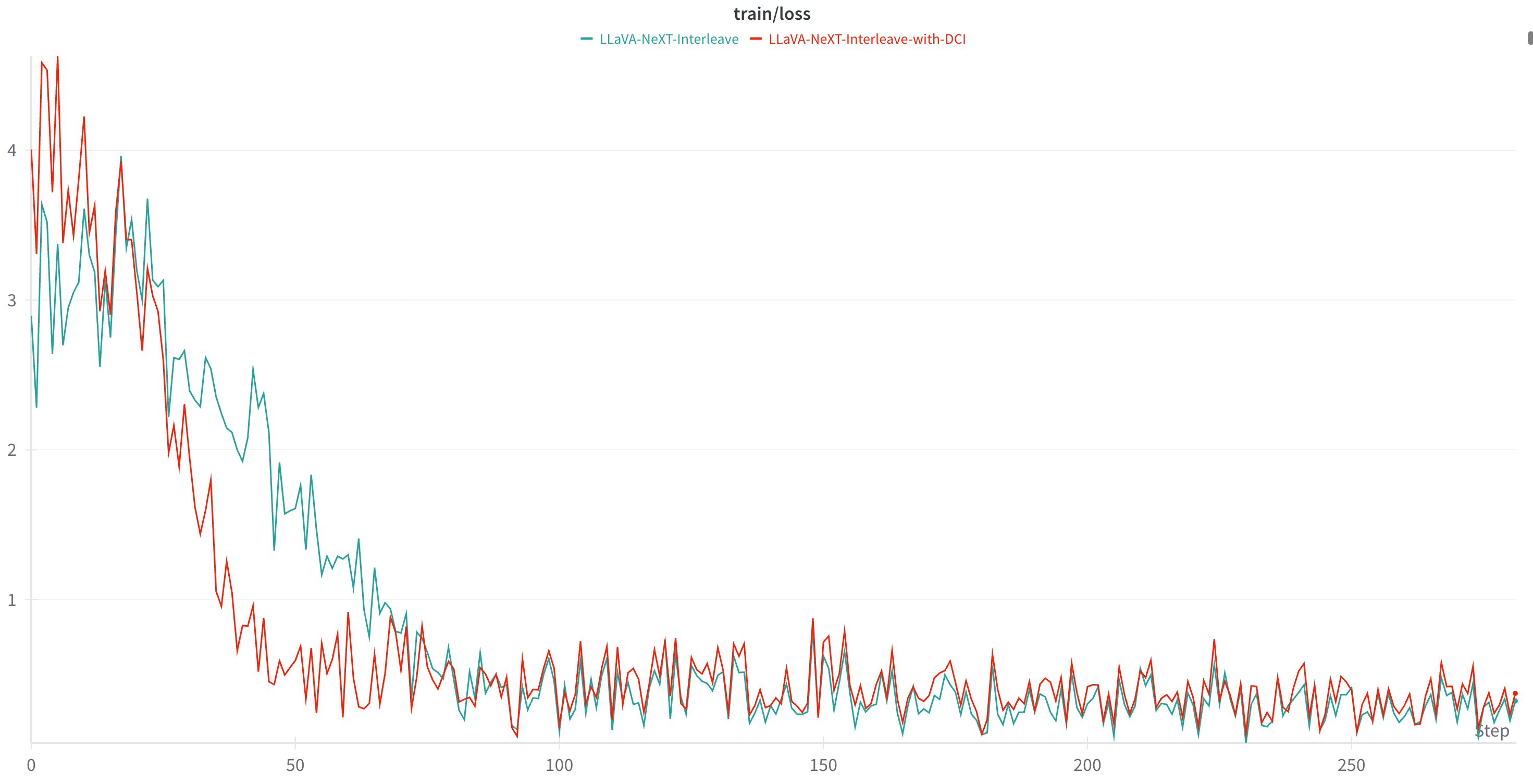}
    \caption{\textbf{Training Loss Comparison between LLaVA-NeXT-Interleave and LLaVA-NeXT-Interleave with DCI.} The DCI-enhanced model converges faster and achieves lower initial training loss, though it exhibits slightly higher variance in later stages.}
    \label{fig:training_loss}
\end{figure}

\paragraph{Training Loss Analysis} The training loss comparison between \textbf{LLaVA-NeXT-Interleave (FT)} and \textbf{FT with Dense Connector Integration (DCI)} as shown in Figure~\ref{fig:training_loss} highlights clear differences. Notably, the DCI-enhanced model \textbf{converges faster} during the early training phase (up to step 60), consistently showing \textbf{lower loss values}. This indicates that DCI facilitates more efficient multimodal fusion and accelerates learning.

However, in the later training stages (after step 100), the DCI variant exhibits \textbf{higher variance}, suggesting a degree of \textbf{optimization instability} likely due to its increased architectural complexity. While both models converge to similar final loss values, DCI’s \textbf{early learning advantage} may benefit tasks requiring rapid adaptation or generalization, albeit with some trade-offs in stability.

\paragraph{Performance on Test set} 

As we gain the Second place for Track A in INOVA Challenge 2025 with the Overall score 74.9, we report our best result on Test set in Table \ref{fig:test_result}. We group each subset by the evaluation metrics to observe the general insights. When considering the \textit{Accuracy} metric, our results behave well with 87.85\% for Reasoning task and 80.52\% for Knowledge Understanding task. The majority of dataset using Accuracy as the evaluation metric gains the 80\%-larger performance, such as The Coherence sets with average 97.13\%. However, with the answer-generating task, the performances are low. For example, on three datasets as Spot-the-Diff, IEdit, and Birds-to-Words, we receive the ROGUE-L scores all under 36.00. This proves that our training strategy is facing a constraint in the free answer task. 

\begin{figure*}[h]
    \centering
    
    % First row
    \begin{subfigure}[t]{0.48\textwidth}
        \centering
        \begin{tabular}{>{\raggedright}p{0.6\textwidth}|c}
            \toprule
            \centering\textbf{DATASET} & \textbf{ROGUE-L} \\
            \midrule
            Spot-the-Diff & 35.58\\ 
            CLEVR-Change & 60.30 \\ 
            IEdit & 32.94 \\ 
            Birds-to-Words & 35.09\\ 
            \midrule
            \textbf{AVERAGE} & \textbf{40.98}\\
            \bottomrule
        \end{tabular}
        \caption{}
    \end{subfigure}
    \hfill
    \begin{subfigure}[t]{0.48\textwidth}
        \centering
        \begin{tabular}{>{\raggedright}p{0.6\textwidth}|c}
            \toprule
            \centering\textbf{DATASET} & \textbf{Accuracy} \\
            \midrule
            nuscenes & 85.80\\
            VISION & 93.60 \\ 
            Fashion200K & 78.40\\
            MIT-States\_PropertyCoherence & 95.00\\ 
            MIT-States\_StateCoherence & 97.80\\ 
            RecipeQA\_ImageCoherence & 98.60 \\ 
            NLVR2 & 88.00\\ 
            VizWiz &65.60 \\ 
            \midrule 
            \textbf{AVERAGE} & \textbf{87.85}\\ 
            \bottomrule
        \end{tabular}
        \caption{}
    \end{subfigure}
    
    \vspace{0.7cm}
    
    % Second row
    \begin{subfigure}[t]{0.48\textwidth}
        \centering
        \begin{tabular}{>{\raggedright}p{0.6\textwidth}|c}
            \toprule
            \centering\textbf{DATASET} & \textbf{Accuracy} \\ \midrule
            SlideVQA & 81.40\\ 
            OCR-VQA & 96.20 \\ 
            WebQA & 86.40 \\ 
            TQA & 85.00\\ 
            MultiModalQA & 53.60\\ \midrule
            % ManyModalQA & \\ \hline
            \textbf{AVERAGE} & \textbf{80.52}\\ \bottomrule
        \end{tabular}
        \caption{}
    \end{subfigure}
    \hfill
    \begin{subfigure}[t]{0.48\textwidth}
        \centering
        \begin{tabular}{>{\raggedright}p{0.6\textwidth}|c}
            \toprule
            \centering\textbf{DATASET} & \textbf{ROGUE-L} \\ \midrule
            MMCoQA & 77.40 \\
            ALFRED & 72.28 \\ \midrule
            \textbf{AVERAGE} & \textbf{74.84} \\ \bottomrule
        \end{tabular}
        \caption{}
    \end{subfigure}
    
    \caption{Performance on Track A's Test set. Multi-Image Reasoning Task result contains 2 tables (a) and (b), while table (c) and (d) represents for the result of Document and Knowledge-Based Understanding task, and Interactive Multi-Modal Communication task respectively.}
    \label{fig:test_result}
\end{figure*}

%% file: section/4-conclusion.tex
\section{Conclusion and Limitations}

In this paper, we apply the Dense Channel Integration connector - a plug-and-play technique to a powerful vision language model - LLaVA-NeXT-Interleave. We compare three versions together and conduct experiments to provide different insights. Based on the results, we believe that there are future directions to explore the ability of the plug-and-play technique in Vision Language Models generally and in Interleave Model specifically. 

\emph{Limitations:} As we just pre-train the model on 1 GPU only, it takes us 22 hours for each epoch run. Moreover, we only pretrain with the 7B-parameter version, that we believe it's still more space to surpass our results.